\documentclass{article}

\usepackage{arxiv}

\usepackage[utf8]{inputenc} % allow utf-8 input
\usepackage[T1]{fontenc}    % use 8-bit T1 fonts
\usepackage{hyperref}       % hyperlinks
\usepackage{url}            % simple URL typesetting
\usepackage{booktabs}       % professional-quality tables
\usepackage{amsfonts}       % blackboard math symbols
\usepackage{nicefrac}       % compact symbols for 1/2, etc.
\usepackage{microtype}      % microtypography
\usepackage{lipsum}
\usepackage{natbib}
\usepackage{amsmath}
\usepackage{listings}
\usepackage{wrapfig}
\usepackage{graphicx}
\graphicspath{ {images/} }
\usepackage{caption}
\usepackage{subcaption}
\usepackage{algpseudocode}
\usepackage{algorithm}
\title{Seismic horizon detection with neural networks}

\author{
  Alexander Koryagin \\
  Data Analysis Center\\
  Gazpromneft\\
  Saint Petersburg \\
   \And
  Darima Mylzenova \\
  Data Analysis Center\\
  Gazpromneft\\
  Saint Petersburg \\
  \And
 Roman Khudorozhkov \\
  Data Analysis Center\\
  Gazpromneft\\
  Saint Petersburg \\
  \And
 Sergey Tsimfer\\
  Data Analysis Center\\
  Gazpromneft\\
  Saint Petersburg \\
}

\begin{document}
\maketitle

%%%%%%%%%%%%%%%%%%%%%%%%%%%%%%%%%%%%%%%%%%%%%%%%%%%%%%%%%%%%%%%%%%%%%%%%%%%%%%%%%%%%%%
\begin{abstract}
Over the last few years, Convolutional Neural Networks (CNNs) were successfully adopted in numerous domains to solve various image-related tasks, ranging from simple classification to fine borders annotation. Tracking seismic horizons is no different, and there are a lot of papers proposing the usage of such models to avoid time-consuming hand-picking. Unfortunately, most of them are (i) either trained on synthetic data, which can't fully represent the complexity of subterranean structures, (ii) trained and tested on the same cube, or (iii) lack reproducibility and precise descriptions of the model-building process. With all that in mind, the main contribution of this paper is an open-sourced research of applying binary segmentation approach to the task of horizon detection on multiple real seismic cubes with a focus on inter-cube generalization of the  predictive model.
\end{abstract}

%%%%%%%%%%%%%%%%%%%%%%%%%%%%%%%%%%%%%%%%%%%%%%%%%%%%%%%%%%%%%%%%%%%%%%%%%%%%%%%%%%%%%%
\section{Introduction}

High-quality tracking of subterranean reflections is of utmost importance to the task of seismic interpretation: it influences all of the subsequent processing steps, including velocity model building. Such picking, on the other hand, requires a lot of time of even very qualified experts, and, consequently, a lot of automatic methods were proposed. Recent works \cite{peters, shengrong} propose to use deep learning based approaches.

Unfortunately, most of the proposed researches focuses on single cube: it is split into train/validation sets, and used to both train the model and evaluate its quality. This approach somewhat reflects the activity of seismic specialist: for example, if the train set consists only of multiple slices, uniformly cut from the cube, and the model is used to <<interpolate>> labeled horizon on the rest of the data. On the other hand, splitting the cube into train/validation sets along one plane is unfair: due to slow changes of information inside the volume, we can't use this validation data to evaluate model performance.

Moreover, we want to use our approach on completely new seismic cubes: to this end, our train set must consist of multiple cubes. Training such models is paired with more obstacles due to varying equipment and settings of shooting, and this raises the bar for the amount of information needed to create a model that can work well on unseen data even higher. Note that we can't use synthetic data to enlarge our dataset: at the moment, we can't plausibly replicate complex subterranean structures as we are unable to reliably simulate time-lasting processes of earth shaping.

%It is well known, that neural networks are data hungry: in classic object recognition tasks size of datasets is at the scales of multiple terabytes. Moreover, information in the train set must be diverse, and this diversity is crucial for generalization of predictive models. Unfortunately, while seismic cubes are big in size, the information is changing very slowly along any of the dimensions: adjacent inlines/cross-lines contain almost identical information. This heavily reduces the seeming diversity and hampers pattern learning. Furthermore, working with multiple seismic cubes at once is paired with more obstacles due to varying equipment and settings of shooting. All this raises the bar for amount of information needed to create a model that can work well on unseen data.

This work is organized as follows: first of all, we describe our data at hand both qualitatively and quantitatively; then we rigorously define task of horizon detection and our approach to it; after that, we strictly define our research on inter-cube generalization and report its results; finally, we discuss shortcomings of the task formulation, as well as propose new method of tracking seismic reflection.

%%%%%%%%%%%%%%%%%%%%%%%%%%%%%%%%%%%%%%%%%%%%%%%%%%%%%%%%%%%%%%%%%%%%%%%%%%%%%%%%%%%%%%
\section{Previous work}

We rely on our framework \textsc{SeismiQB}, which was presented in [CITE]. To briefly recap, it allows to store seismic data in a fast data format together with attached horizons (either hand-labeled or any other), cut crops of desired shape from it and apply various transformations to the values. This library also allows to easily define even the most sophisticated neural networks with just a few lines of code, and there are also implementations for various popular architectures ranging from simple \textsc{ResNet} \cite{resnet} and \textsc{UNet}-like \cite{unet} models to state-of-the-art \textsc{EfficientNet} \cite{efficientnet} and \textsc{DeepLab} \cite{deeplab}. 

\begin{figure}
    \centering
    \begin{minipage}{.5\textwidth}
      \centering
      \includegraphics[width=8cm, height=8cm]{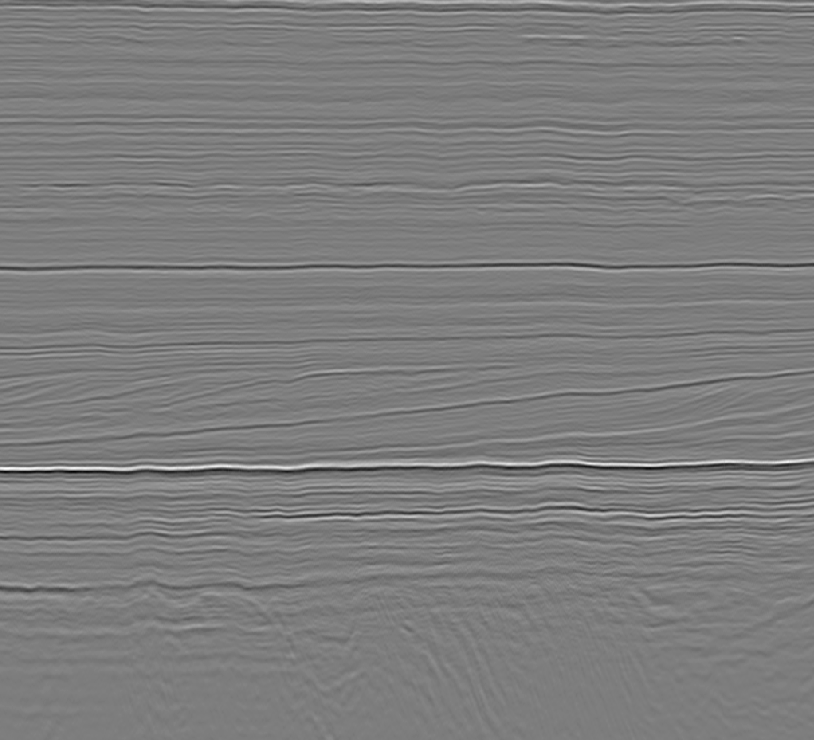}
      \captionof{figure}{Example of data from Cube 3}
      \label{fig:cube_3}
    \end{minipage}%
    \begin{minipage}{.5\textwidth}
      \centering
      \includegraphics[width=8cm, height=8cm]{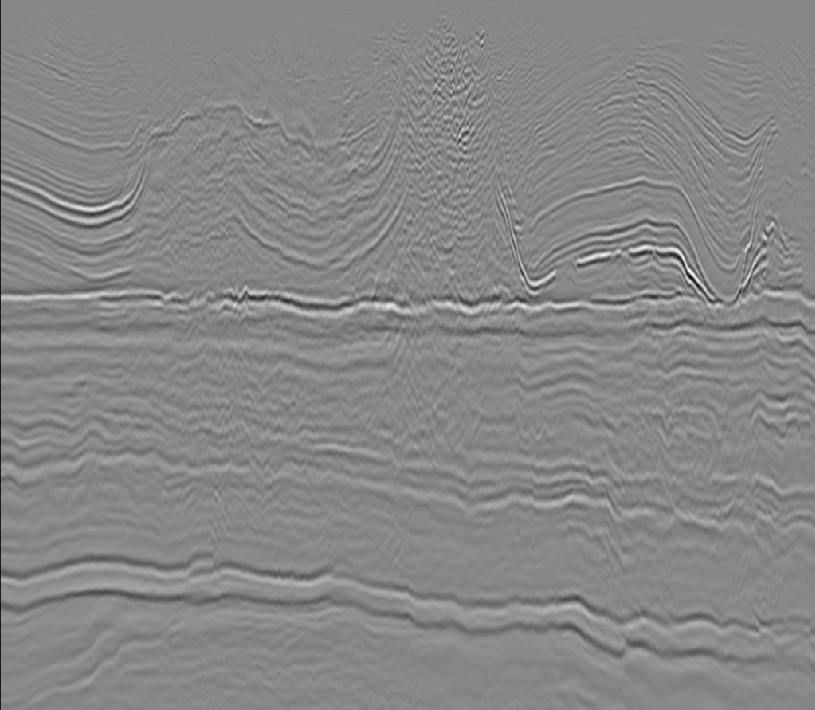}
      \captionof{figure}{Example of data from Cube 4}
      \label{fig:cube_4}
    \end{minipage}
\end{figure}

Table \ref{table:dataset} provides detailed information about four of the (anonymized) cubes in our dataset. Images \ref{fig:cube_3}, \ref{fig:cube_4} illustrate completely different inner structure of seismic data, which is a consequence of both varying equipment of shooting and different subterranean patterns in distant locations. We can see, that \texttt{Cube\,3} is very plain and almost linear, while \texttt{Cube\,4} contains sharp turns and fissures, so we expect varying performance of the neural networks on them. Each of the cubes is paired with a number (from four to nine) of hand-labeled horizons which have diverse nature: some of them track the most noticeable reflections, some of them mark boundaries between interlayers, few of them follow fissures and cracks. Worse yet, they track different phase: some of them are at maximas of amplitudes, some of them are at minimas, while most of human-picked horizons have varying phase over the domain. Quality evaluation of tracked reflections must be further explored: training models on bad data is obviously detrimental to its predictive ability. Information about the amount of labeled horizons for each cube is also presented in Table \ref{table:dataset}.

\begin{table}[h]
\centering
 \begin{tabular}{||c | c | c | c||} 
 \hline
 \textbf{Cube alias} & \textbf{Size, GB} & \textbf{Shape (inlines, cross-line, depth} & \textbf{Number of labeled horizons} \\ [0.5ex] 
 \hline
 \textbf{Cube 1} & 21 & 2563, 1409, 1501 & 9 \\ 
 \hline
 \textbf{Cube 2} & 2.8 & 418, 869, 2001 & 5 \\
 \hline
 \textbf{Cube 3} & 54 & 2737, 2599, 2001 & 4 \\
 \hline
 \textbf{Cube 4} & 8.2 & 1472, 1380, 1071 & 8 \\
 \hline
\end{tabular}
\caption{Results for models that are trained and tested on the same cube\vspace{2em}}
\label{table:dataset}
\end{table}

\vspace{-3em}
\textsc{SeismiQB}'s capability to transform values in the cube is crucial: as we know, cubes contain excessive information with the relation to the task of tracking reflections. Thus, we can use a wealth of functions to transform the space of $(x, y, t)$ to the space of $(x, y, \omega)$ with $\omega$ being an arbitrary hand-developed feature like frequency, sign, phase, etc, in order to represent information in a more suitable way. This new data can be used as a complementary one, or replace original image entirely. In section [REF SECTION] we investigate some of the transforms and their influence on the quality of the predictive model.

%%%%%%%%%%%%%%%%%%%%%%%%%%%%%%%%%%%%%%%%%%%%%%%%%%%%%%%%%%%%%%%%%%%%%%%%%%%%%%%%%%%%%%
\section{Formal task description}
There are multiple approaches to solve the task in hand: we can cast it as a regression problem, or as binary or multiclass segmentation. In this paper, we use the binary segmentation method. We use $N_x, N_y, N_t$ as  sizes of crop along each respective dimension, and $n$ is reserved for the number of horizons.

Despite every crop being a 3D entity cut from the original cube, we don't use 3D (volumetric) convolutions due to no metric gains and prohibitive computational costs of such neural networks. Consequently, our models perceive input data as 2D entity with channels, and uses 2D (spatial) convolutions to process it. In order to cast the task into a binary segmentation one, model should predict 3D array of the same shape, as the input data, with 1's corresponding to horizons and 0's to background. Thus, CNN should learn mapping from $\mathcal{R}^{N_x  \times N_y \times N_t}$ on itself, which is commonly achieved by an encoder-decoder architectures. We evaluated multiple models in such fashion, ranging from simple \textsc{UNet} \cite{unet} to the state-of-the-art \textsc{DeepLab} \cite{deeplab}. We also want our model to look at the image from a side view: to this end, we use any of the spatial axis (cross-line or inline) as the channel one.

Due to heavy class imbalance (volume of horizons is extremely small compared to the volume of the whole image), we use Dice coefficient as the loss function. Note that we can artificially increase the volume of horizon class by thickening the labeled surface: that helps model to distinguish classes. On the other hand, that hampers its ability to finely locate reflections, which is undesirable, and it also makes some of the closer horizon to overlap with one another, which is absolutely unacceptable.

%%%%%%%%%%%%%%%%%%%%%%%%%%%%%%%%%%%%%%%%%%%%%%%%%%%%%%%%%%%%%%%%%%%%%%%%%%%%%%%%%%%%%%
\section{Results}
\subsection{Train protocol}
We train models on batches of 64 randomly generated crops for 1000 iterations by Adam optimizer with default parameters, augmented by inverse-time learning rate decay schedule, which takes approximately one hour to finish. Crops are scaled to a $[0, 1]$ range via min-max scaling in order to have the same range of values in every cube. Also, at train time we augment crops with various distortions: additive and multiplicative noises, affine and perspective transformations, cutout \cite{cutout} and elastic transform \cite{elastic}. This process is aimed at making model robust to imperfections of the data and to make training examples more diverse.

The shape of the input crops deserves special attention. First observation is that cube changes very slowly along its spatial dimensions: therefore, there is no need to make it big in both cross-lines and inlines, and keeping one of them big enough should suffice. Secondly, shape can either be fixed during the whole time of training and inference or dynamically change at training time. We found that randomly generating crop shape from tenth of the cube to its half helps inter-cube generalization, but takes much more time for training (two or three times longer) and prone to unstable results.

\subsection{Train setups}
In order to exhaustively study inter-cube generalization, we've prepared multiple combinations of train and validation setups. That allows us to thoroughly examine the problem and make definitive conclusions about the obtained results. To evaluate the quality of the predicted horizons, we use multiple metrics, including the percentage of covered area (compared to the ground truth), mean error ($l_1$-difference) and percentage of area of predicted horizon inside 5-ms window of the hand-labeled one.

\begin{wrapfigure}{r}{0.3\textwidth}
  \vspace{-13pt}
  \centering
  \includegraphics[width=.3\textwidth]{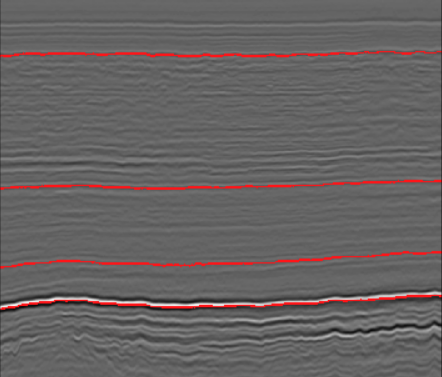}
  \captionof{figure}{Example of model labelling on unseen data}
  \vspace{-20pt}
  \label{fig:unseen}
\end{wrapfigure}

First of all, we train and test neural network on the same cube. We use only every 200-th inline during train, totaling in no more than 15 slides for each individual cube as a train data. This setting corresponds to a situation, where seismic specialist labels sought-for horizon only on every 200-th slide and wants its automatic prolongation on the whole cube. Due to slow change of the cube along its spatial axis, we anticipate good performance in this task. Results, presented in Table \ref{table:first}, are consistent with our expectations: predicted horizons cover most of the needed area while staying close to the hand-labeled ones. Qualitative estimation shows that prediction makes sense from a geological point of view: it follows the same phase, having discontinuities only in the locations of fissures or sharp faults.

\begin{table}[h]
\centering
 \begin{tabular}{||c | c | c | c||} 
 \hline
 \textbf{Train/test cube} & \textbf{Area, \%} & \textbf{Mean error, ms} & \textbf{Area in 5ms window, \%} \\ [0.5ex] 
 \hline
 \textbf{Cube 1} & 90, 91, 86 & 1.5, 2.1, 4.4 & 96, 94, 85 \\ 
 \hline
 \textbf{Cube 2} & 100, 98, 95 & 2.4, 2.1, 2.8 & 96, 99.9, 96.4 \\
 \hline
 \textbf{Cube 3} & 92, 92, 92 & 1.8, 1.9, 2.5 & 99.8, 99.5, 91.6 \\
 \hline
 \textbf{Cube 4} & 73, 83, 84 & 3, 3.6, 4.3 & 89, 86, 84 \\
 \hline
\end{tabular}
\caption{Results for models that are trained and tested on the same cube}
\label{table:first}
\end{table}

Model, trained on one cube, hardly works on the unseen data. That can be attributed to varying subterranean structure, as well as to the unequal values inside individual cubes: despite scaling values to a $[0, 1]$ range, the exact distribution can diverse. Thus, we need to train the model on multiple cubes in order to work in other regions.

Next, we make model learn from two cubes with validation of the results on the third. Note that, due to small size, we exclude \texttt{Cube\,2} from this part of research: it is just not big enough. All in all, we trained three models, with results shown in Table \ref{table:pairs}. We can easily spot a glaring problem here: predictions cover only a fraction of the spatial size. Manual check allows us to conclude that the model works somewhat well (detected horizons are in 5ms window) in places where the texture of the validation data is similar to the learned one, and in order to generalize better we must provide more diverse seismic information at train time.

Our last model, trained on three cubes and tested on \texttt{Cube\,2}, confirms this statement, achieving both good coverage and mean error results, presented in Table \ref{table:last}. An example of a model working on unseen data can be seen at Image \ref{fig:unseen}.

\vspace{2.5em}

\begin{table}[h]
\centering
 \begin{tabular}{||c | c | c | c | c||} 
 \hline
 \textbf{Train cube} & \textbf{Test cube} & \textbf{Area, \%} & \textbf{Mean error, ms} & \textbf{Area in 5ms window, \%} \\ [0.5ex] 
 \hline
 \textbf{Cube 3, Cube 4} & \textbf{Cube 1} & 5 & 0.85 & 94 \\ 
 \hline
 \textbf{Cube 1, Cube 4} & \textbf{Cube 3} & 45 & 1 & 97 \\
 \hline
 \textbf{Cube 1, Cube 3} & \textbf{Cube 4} & 25 & 50 & 0 \\
 \hline
\end{tabular}
\caption{Results for models that are trained on two cubes and tested on the unseen data}
\label{table:pairs}
\end{table}

\begin{table}[h!]
\centering
 \begin{tabular}{||c | c | c | c | c||} 
 \hline
 \textbf{Train cube} & \textbf{Test cube} & \textbf{Area, \%} & \textbf{Mean error, ms} & \textbf{Area in 5ms window, \%} \\ [0.5ex] 
 \hline
 \textbf{Cube 1, Cube 3, Cube 4} & \textbf{Cube 2} & 98 & 0.81 & 98.9 \\ 
 \hline
\end{tabular}
\caption{Results for model that is trained on three cubes and tested on the unseen data}
\label{table:last}
\end{table}

%%%%%%%%%%%%%%%%%%%%%%%%%%%%%%%%%%%%%%%%%%%%%%%%%%%%%%%%%%%%%%%%%%%%%%%%%%%%%%%%%%%%%%
\section{Discussion}
Despite accomplishing acceptable results, there is a major issue. The task in hand is ill-defined: it is unclear, which of the horizons do we want to detect, how many of them, with which rules of tracking (e.g. on which phase the horizon should be). Moreover, we have virtually no control at the model's behavior on unseen data: we can't choose which horizon to track. That contradicts with the very nature of the problem: in most situations, we want to know the location of a particular chosen horizon.

There is a simple way of attacking this problem: just get more train labels, one way or the other. That would require significant human resources, yet would not solve the issue completely: we would make model to detect more horizons, but that does not eliminate situations where we need to track the one, not present in predictions. This is a simple, but a rather unscalable approach.

A better way is to communicate prior information about horizon location to the model, at the same time making it follow only one reflection at a time. Doing so requires significant enhancement of our framework, but appears to be the best way to tackle the issue. Introducing a mechanism of hand-selecting horizon to track would both make results more interpretable and appealing. 

%%%%%%%%%%%%%%%%%%%%%%%%%%%%%%%%%%%%%%%%%%%%%%%%%%%%%%%%%%%%%%%%%%%%%%%%%%%%%%%%%%%%%%
\section{Conclusion}
In this paper, we systematically study one of the approaches to seismic horizon detection. With multiple carefully designed settings we identify both situations where deep learning based methods can work well, as well as harder tasks that are yet to be successfully tackled by neural networks; in each of the settings we show both quantitive and qualitative results. We also demonstrate crucial problems with the task itself and propose to move away from unsupervised horizon detection (in a sense that we don't control locations of the predicted horizons) to a supervised one.

\bibliographystyle{plainnat}
\bibliography{references}

\begin{thebibliography}{8}
\providecommand{\natexlab}[1]{#1}
\providecommand{\url}[1]{\texttt{#1}}
\expandafter\ifx\csname urlstyle\endcsname\relax
  \providecommand{\doi}[1]{doi: #1}\else
  \providecommand{\doi}{doi: \begingroup \urlstyle{rm}\Url}\fi

\bibitem[Chen et~al.(2018)Chen, Zhu, Papandreou, Schroff, and Adam]{deeplab}
Liang{-}Chieh Chen, Yukun Zhu, George Papandreou, Florian Schroff, and Hartwig
  Adam.
\newblock Encoder-decoder with atrous separable convolution for semantic image
  segmentation.
\newblock \emph{CoRR}, abs/1802.02611, 2018.
\newblock URL \url{http://arxiv.org/abs/1802.02611}.

\bibitem[Devries and Taylor(2017)]{cutout}
Terrance Devries and Graham~W. Taylor.
\newblock Improved regularization of convolutional neural networks with cutout.
\newblock \emph{CoRR}, abs/1708.04552, 2017.
\newblock URL \url{http://arxiv.org/abs/1708.04552}.

\bibitem[He et~al.(2015)He, Zhang, Ren, and Sun]{resnet}
Kaiming He, Xiangyu Zhang, Shaoqing Ren, and Jian Sun.
\newblock Deep residual learning for image recognition.
\newblock \emph{CoRR}, abs/1512.03385, 2015.
\newblock URL \url{http://arxiv.org/abs/1512.03385}.

\bibitem[Li et~al.(2019)Li, Yang, Sun, and Zhang]{shengrong}
Shengrong Li, Changchun Yang, Hui Sun, and Hao Zhang.
\newblock {Seismic fault detection using an encoder–decoder convolutional
  neural network with a small training set}.
\newblock \emph{Journal of Geophysics and Engineering}, 16\penalty0
  (1):\penalty0 175--189, 03 2019.
\newblock ISSN 1742-2132.
\newblock \doi{10.1093/jge/gxy015}.
\newblock URL \url{https://doi.org/10.1093/jge/gxy015}.

\bibitem[{Peters} et~al.(2018){Peters}, {Granek}, and {Haber}]{peters}
Bas {Peters}, Justin {Granek}, and Eldad {Haber}.
\newblock {Multi-resolution neural networks for tracking seismic horizons from
  few training images}.
\newblock \emph{arXiv e-prints}, art. arXiv:1812.11092, Dec 2018.

\bibitem[Ronneberger et~al.(2015)Ronneberger, Fischer, and Brox]{unet}
Olaf Ronneberger, Philipp Fischer, and Thomas Brox.
\newblock U-net: Convolutional networks for biomedical image segmentation.
\newblock \emph{CoRR}, abs/1505.04597, 2015.
\newblock URL \url{http://arxiv.org/abs/1505.04597}.

\bibitem[{Simard} et~al.(2003){Simard}, {Steinkraus}, and {Platt}]{elastic}
P.~Y. {Simard}, D.~{Steinkraus}, and J.~C. {Platt}.
\newblock Best practices for convolutional neural networks applied to visual
  document analysis.
\newblock pages 958--963, Aug 2003.
\newblock \doi{10.1109/ICDAR.2003.1227801}.

\bibitem[Tan and Le(2019)]{efficientnet}
Mingxing Tan and Quoc~V. Le.
\newblock Efficientnet: Rethinking model scaling for convolutional neural
  networks.
\newblock \emph{CoRR}, abs/1905.11946, 2019.
\newblock URL \url{http://arxiv.org/abs/1905.11946}.

\end{thebibliography}

\end{document}